\documentclass{article} 
\pdfoutput=1
\usepackage{iclr2019_conference,times}

\usepackage{amsmath,amsfonts,bm}

\def\figref#1{Fig.~\ref{#1}}
\def\Figref#1{Fig.~\ref{#1}}

\def\secref#1{Sec.~\ref{#1}}
\def\Secref#1{Sec.~\ref{#1}}

\def\eqref#1{equation~\ref{#1}}

\def\1{\bm{1}}

\DeclareMathAlphabet{\mathsfit}{\encodingdefault}{\sfdefault}{m}{sl}
\SetMathAlphabet{\mathsfit}{bold}{\encodingdefault}{\sfdefault}{bx}{n}

\usepackage[pagebackref=true,breaklinks=true,colorlinks,bookmarks=false,citecolor=blue]{hyperref}
\usepackage{hyperref}
\usepackage{url}
\usepackage{lib}
\usepackage{siunitx}
\usepackage{xcolor}
\usepackage{wrapfig}
\usepackage{graphicx}
\usepackage[export]{adjustbox}
\usepackage{float}
\usepackage{booktabs}
\usepackage{xspace}
\usepackage{enumitem}
\usepackage{blindtext}
\usepackage{subcaption}
\usepackage[title]{appendix}
\usepackage[font=footnotesize,labelfont=bf]{caption}
\title{Learning Exploration Policies for Navigation}

\author{Tao Chen$^1$ $\quad$ Saurabh Gupta$^2$ $\quad$ Abhinav Gupta$^{1,2}$ \\
$^1$Carnegie Mellon University \\
$^2$Facebook AI Research \\
}

\newcommand{\rgbd}{RGB-D\xspace}
\newcommand{\rgb}{RGB\xspace}

\newcommand{\spl}{SPL\xspace}
\newcommand{\house}{House 3D\xspace}
\newcommand{\embodiedqa}{EmbodiedQA\xspace}

\newcommand{\insertW}[2]{\IfFileExists{#2}{\includegraphics[width=#1\textwidth]{#2}}{\includegraphics[width=#1\textwidth]{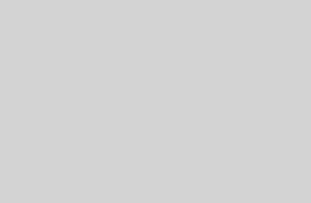}}}

\iclrfinalcopy 
\begin{document}

\maketitle

\begin{abstract}
Numerous past works have tackled the problem of task-driven navigation. But, how to effectively explore a new environment to enable a variety of down-stream tasks has received much less attention. In this work, we study how agents can autonomously explore realistic and complex 3D environments without the context of task-rewards. We propose a learning-based approach and investigate different policy architectures, reward functions, and training paradigms. We find that use of policies with spatial memory that are bootstrapped with imitation learning and finally finetuned with coverage rewards derived purely from on-board sensors can be effective at exploring novel environments. We show that our learned exploration policies can explore better than classical approaches based on geometry alone and generic learning-based exploration techniques. Finally, we also show how such task-agnostic exploration can be used for down-stream tasks. Code and Videos are available at: \url{https://sites.google.com/view/exploration-for-nav/}.
\end{abstract}

\section{Introduction}
Imagine your first day at a new workplace. If you are like most people, the first task you set for yourself is to become familiar with the office so that the next day when you have to attend meetings and perform tasks, you can navigate efficiently and seamlessly. To achieve that goal, you explore your office without the task context of target locations you have to reach and build a generic understanding of space. This step of task-independent exploration is quite critical yet often ignored in current approaches for navigation.

When it comes to navigation, currently there are two 
paradigms: (a) geometric reconstruction and path-planning based 
approaches~\citep{hartley2003multiple, thrun2005probabilistic, lavalle2006planning}, 
and (b) learning-based approaches~\citep{mirowski2016learning, gupta2017cognitive, savinov2018semi, zhu2016target}. 
SLAM-based approaches, first build a map and then use localization and 
path planning for navigation. In doing this, one interesting question that is often 
overlooked is: How does one build a map? How should we explore the 
environment to build this map? Current approaches either use a human 
operator to control the robot for building the map (\eg \cite{thrun1999minerva}),
or use heuristics such as frontier-based exploration~\citep{yamauchi1997frontier}. 
On the other hand for approaches that use learning, most only learn 
policies for specific tasks~\citep{zhu2016target, mirowski2016learning, gupta2017cognitive},
or assume environments have already been explored~\citep{savinov2018semi}.
Moreover, in context of such learning based approaches, the question of exploration
is not only important at test time, but also at train time. And once again, this question 
is largely ignored. Current approaches either use sample inefficient 
random exploration or make impractical assumptions about full map 
availability for generating supervision from optimal trajectories.

Thus, a big bottleneck for both these navigation paradigms 
is an exploration policy: a task-agnostic policy that explores the 
environment to either build a map or sample trajectories for learning 
a navigation policy in a sample-efficient manner. But how do we learn 
this task-independent policy? What should be the reward for such a policy? 
First possible way is to not use learning and use heuristic based 
approaches~\citep{yamauchi1997frontier}. However, there are four issues with 
non-learning based approaches: (a) these approaches are brittle and 
fail when there is noise in ego-estimation or localization; (b) they make 
strong assumptions about free-space/collisions and fail to generalize 
when navigation requires interactions such as opening doors \etc; (c) they 
fail to capture semantic priors that can reduce search-space significantly;
and (d) they heavily rely on specialized sensors such as range scanners.
Another possibility is to learn exploration policies on training environments. 
One way is to use reinforcement learning (RL) with intrinsic rewards. 
Examples of intrinsic rewards can be ``curiosity'' where prediction error 
is used as reward signal or ``diversity'' which discourages the agent 
from revisiting the same states. While this seems like an effective reward, 
such approaches are still sample inefficient due to blackbox reward functions 
that can't be differentiated to compute gradients effectively. 
So, what would be an effective way to learn exploration policies for navigation? 

In this paper, we propose an approach for learning policies for exploration
for navigation. We explore this problem from multiple perspectives: 
(a) architectural design, (b) reward function design, and (c) reward optimization. 
Specifically, 
from perspective of reward function and optimization, we take the 
alternative paradigm and use supervision from human explorations in 
conjunction with intrinsic rewards.
We notice that bootstrapping of learning from small amount of 
human supervision aids learning of semantics (\eg doors are pathways). 
It also provides a good initialization for learning using intrinsic rewards. 
From the perspective of architecture design, we explore how to use 
3D information and use it efficiently while doing exploration. We study proxy rewards that characterize coverage and demonstrate how 
this reward outperforms other rewards such as curiosity.
Finally, we show how experience gathered from our learned exploration 
policies improves performance at down-stream navigation tasks.

\section{Related Work}
Our work on learning exploration policies for navigation 
in real world scenes is related to active SLAM in 
classical robotics, and intrinsic rewards based exploration
in reinforcement learning. As we study the problem of 
navigation, we also draw upon recent efforts that use learning 
for this problem. We survey related efforts in these three directions.

\textbf{Navigation in Classical Robotics.}
Classical approaches to navigation operate by building a map of the environment, localizing the agent in this map, and then planning paths to convey the agent to desired target locations. Consequently, the problems of mapping, localization and path-planning have been very thoroughly studied \citep{hartley2003multiple, thrun2005probabilistic, lavalle2006planning}. However, most of this research starts from a human-operated traversal of the environment, and falls under the purview of passive SLAM. Active SLAM, or how to automatically traverse a new environment for building spatial representations is much less studied. \cite{cadena2016past} present an excellent review of active SLAM literature, we summarize some key efforts here.  Past works have formulated active SLAM as Partially Observable Markov Decision Processes (POMDPs) \citep{martinez2009bayesian}, or as choosing actions that reduce uncertainty in the estimated map \citep{carrillo2012comparison}. While these formulations enable theoretical analysis, they crucially rely on sensors to build maps and localize. Thus, such approaches are highly susceptible to measurement noise. Additionally, such methods treat exploration purely as a geometry problem, and entirely ignore semantic cues for exploration such as doors.

\textbf{Learning for Navigation.}
In order to leverage such semantic cues for navigation, recent works have formulated navigation as a learning problem \citep{zhu2016target, gupta2017cognitive, mirowski2016learning, savinov2018semi}. A number of design choices have been investigated. For example, these works have investigated different policy architectures for representing space: \citet{zhu2016target} use feed-forward networks, \citet{mirowski2016learning} use vanilla neural network memory, \citet{gupta2017cognitive} use spatial memory and planning modules, and \citet{savinov2018semi} use semi-parametric topological memory. Different training paradigms have also been explored: \citet{gupta2017cognitive} learn to imitate behavior of an optimal expert, \citet{mirowski2016learning} and \cite{zhu2016target} use extrinsic reward based reinforcement learning, \cite{pathakICLR18zeroshot} learn an inverse dynamics model on the demonstrated trajectory way-points from the expert, while \citet{savinov2018semi} use self-supervision. In our work here, we build on insights from these past works. Our policy architecture and reward definition use spatial memory to achieve long-horizon exploration, and we imitate human exploration demonstrations to boot-strap policy learning. However, in crucial distinction, instead of studying goal-directed navigation (either in the form of going to a particular goal location, or object of interest), we study the problem of autonomous exploration of novel environments in a task-agnostic manner. In doing so, unlike past works, we do not assume access to human demonstrations in the given novel test environment like \cite{savinov2018semi}, nor do we assume availability of millions of samples of experience or reward signals in the novel test environment like \citet{zhu2016target} or \citet{mirowski2016learning}. Moreover, we do not depend on extrinsically defined reward signals for training. We derive them using on-board sensors. This makes our proposed approach amenable to real world deployment. Learning based approaches~\citep{gandhi2017learning, sadeghi2016cad2rl} have also been used to learn low-level collision avoidance policies. While they do not use task-context, they learn to move towards open space, without the intent of exploring the whole environment. \citet{zhangneural} uses a differentiable map structure to mimic the SLAM techniques. Such works are orthogonal to our effort on exploration as our exploration policy can benefit from their learned maps instead of only using reconstructed occupancy map.

\textbf{Exploration for Navigation.}
A notable few past works share a similar line of thought, 
and investigate exploration in context of reinforcement learning 
\citep{schmidhuber1991possibility, stadie2015incentivizing, 
pathak2017curiosity, fu2017ex2, lopes2012exploration, chentanez2005intrinsically}. 
These works design intrinsic reward functions to capture novelty of states or state-action transitions. Exploration policies are then learned by optimizing these reward functions using reinforcement learning. Our work is most similar to these works. However, we focus on this problem in the specific context of navigation in complex and realistic 3D environments, and propose specialized policy architectures and intrinsic reward functions. We experimentally demonstrate that these specializations improve performance, and how learning based exploration techniques may not be too far from real world deployment. \cite{jayaraman2018learning} use a related reward function (pixel reconstruction) 
to learn policies to look around (and subsequently solve tasks). However, 
they do it in context of 360$^\circ$ images, and their precise reward 
can't be estimated intrinsically. \citet{xu2017autonomous} generate smooth movement path for high-quality camera scan by using time-varying tensor fields. \citet{bai2016information} propose an information-theoretic exploration method using Gaussian Process regression and show experiments on simplistic map environments. \citet{kollar2008trajectory} assume access to the ground-truth map and learn an optimized trajectory that maximizes the accuracy of the SLAM-derived map. In contrast, our learning policy directly tells the action that the agent should take next and estimates the map on the fly. 

\textbf{System Identification.} Finally, the general idea of exploration prior to goal-driven behavior, is also related to the classical idea of system identification \citep{ljung1987system}. Recent interest in this idea with end-to-end learning \citep{yu2017preparing, duan2016rl, mishra2018simple} has been shown to successfully adapt to novel mazes. In comparison, we tackle navigation in complex and realistic 3D environments, with very long spatio-temporal dependencies.

\section{Approach}
\begin{figure}
\centering
\vspace{-40pt}
\insertW{0.9}{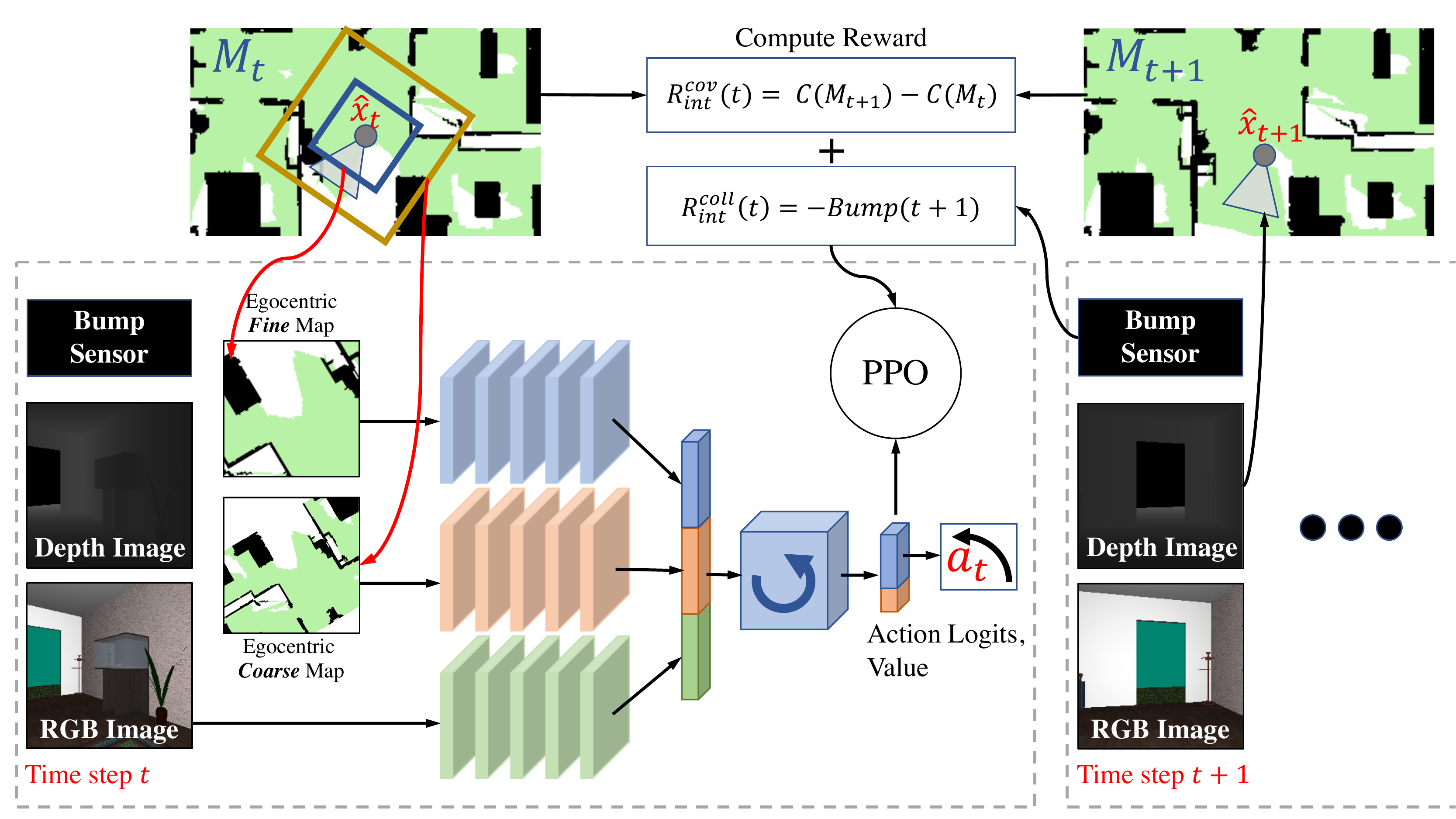}
\caption{\textbf{Policy and Training Architecture:}
Our scheme for learning exploration policies for navigation.
We assume a mobile robot with an \rgbd camera and a bump sensor.
We maintain a running estimate of its position $\hat{x}_t$. These position
estimates are used to stitch together an approximate map of the environment
using the depth images. The exploration policy $\pi_e$ is a recurrent network
takes egocentric crops of this map at two scales, and the current \rgb image 
as input. The policy is trained using an intrinsic coverage reward that 
is computed from the approximate map and a collision penalty obtained 
from the bump sensor.}
\label{fig:policy}
\vspace{-12pt}
\end{figure}

Let us consider a mobile agent that is equipped with an \rgbd
camera and that can execute basic movement macro-actions. This agent has been 
dropped into a novel environment that it has never been in before.  
We want to learn an exploration policy 
$\pi_e$ that enables this agent to efficiently explore this new 
environment. A successful exploration of the new environment
should enable the agent to accomplish down-stream tasks in this 
new environment efficiently.

We formulate estimation of $\pi_e$ as a learning problem. 
We design a reward function that is estimated entirely using on-board sensors,
and learn $\pi_e$ using RL. Crucially, $\pi_e$ is learned on a set of environments
$\mathcal{E}_{train}$, and tested on a held-out set of environments
$\mathcal{E}_{test}$, \ie, $\mathcal{E}_{train} \bigcap \mathcal{E}_{test} = \phi$.

The key novelties in our work are: a) the design of the policy $\pi_e$,
b) the design of the reward function $R$, and c) use of human demonstrations
to speed up training of $\pi_e$. The design of the policy and the reward function
depend on a rudimentary map of the environment that we maintain over time.
We first describe the map construction procedure, and then detail the aforementioned aspects.

\subsection{Map Construction}

\label{sec:make-map}
As the agent moves around, it uses its depth camera to build and update a map of the world around it. This is done by maintaining an estimate of the agent's location over time and projecting 3D points observed in the depth image into an allocentric map of the environment. 

More specifically, let us assume that the agent's estimate of its current location at time step $t$ is $\hat{x_t}$, and that the agent starts from origin (\ie $\hat{x}_0=\bm{0}$). When the agent  executes an action $a_t$ at time step $t$, it updates its position estimate through a known transition function $f$, \ie, $\hat{x}_{t+1} = f(\hat{x}_{t}, a_t)$.

The depth image observation at time step $t+1$, $D_{t+1}$ is back-projected into 3D points using known camera intrinsic parameters. These 3D points are transformed using the estimate $\hat{x}_{t+1}$, and projected down to generate a 2D map of the environment (that tracks what parts of space are known to be traversable or known to be non-traversable or unknown). Note that,
because of slippage in wheels \etc, $\hat{x}_{t+1}$ may not be the same as the true relative location of the agent from start of episode $x_{t+1}$. This leads to aliasing in the generated map. We do not use expensive non-linear optimization (bundle adjustment) to improve our estimate $\hat{x}_{t+1}$, but instead rely on learning to provide robustness against this mis-alignment. We use the new back-projected 3D points to update the current map $M_{t}$ and to obtain an updated map $M_{t+1}$.

\subsection{Policy Architecture}

Before we describe our policy architecture, let's define what are the 
features that a good exploration policy needs: 
(a) good exploration of a new environment requires an agent to meaningfully move around by detecting and avoiding obstacles; 
(b) good policy also requires the agent to identify semantic cues such as doors  that may facilitate exploration; 
(c) finally, it requires the agent to keep track of what parts of the environment have or have not been explored, and to estimate how to get to parts of the environment that may not have been explored. 

This motivates our policy architecture. We fuse information from 
\rgb image observations and occupancy grid-based maps. 
Information from the \rgb image allows recognition of useful semantic cues.
While information from the occupancy maps allows the agent 
to keep track of parts of the environment that have or have not been explored and 
to plan paths to unexplored regions without bumping into obstacles.

The policy architecture is shown in \figref{fig:policy}.  
We describe it in detail here:
\begin{enumerate}[leftmargin=*]
\item \textbf{Information from \rgb images}: \rgb images are processed through 
    a CNN. We use a ResNet-18 CNN that has been pre-trained on the ImageNet
    classification task, and can identify semantic concepts in images. 
\item \textbf{Information from Occupancy Maps}: 
    We derive an occupancy map from past observations (as described in 
    \secref{sec:make-map}), and use it with a ResNet-18 CNN to extract features. 
    To simplify learning, we do not
    use the allocentric map, but transform it into an egocentric map using the 
    estimated position $\hat{x}_{t}$ (such that the agent is always at the
    center of the map, facing upwards). This map canonicalization
    aids learning. It allows the CNN to not only detect unexplored space but to also
    locate it with respect to its current location.
    Additionally, we use two such egocentric maps, a coarse map that captures
    information about a $40m \times 40m$ area around the agent, and a detailed map
    that describes a $4m \times 4m$ neighborhood. Some of these design choices 
    were inspired by recent work from \cite{gupta2017cognitive}, 
    \cite{parisotto2018neural} and \cite{henriques2018mapnet}, 
    though we make some simplifications.
\item \textbf{Recurrent Policy}: Information from the \rgb image and maps is
    fused and passed into an RNN. Recurrence can allow the agent to exhibit coherent
    behavior.
\end{enumerate}

\subsection{Coverage Reward}
\label{sec:reward}
We now describe the intrinsic reward function that we use to train 
our exploration policy. We derive this intrinsic rewards from the map
$M_t$, by computing the coverage. Coverage $C(M_t)$ is defined as the
total area in the map that is known to be traversable or known to be non-traversable.
Reward $R^{cov}_{int}(t)$ at time step $t$ is obtained via gain in coverage in the map: 
$C(M_{t+1}) - C(M_{t})$. Intuitively, if the current observation 
adds no obstacles or free-space to the map then it adds no information 
and hence no reward is given. We also use a collision penalty, that
is estimated using the bump sensor, $R^{coll}_{int}(t) = -Bump(t+1)$, where 
$Bump(t+1)$ denotes if a collision occurred while executing action $a_t$.
$R^{cov}_{int}$ and $R^{coll}_{int}$ are combined to obtain the total reward.

\subsection{Training Procedure}
\label{sec:training}
Finally, we describe how we optimize the policy. Navigating in complex
realistic 3D houses, requires long-term coherent behavior over multiple time steps,
such as exiting a room, going through doors, going down hallways. 
Such long-term behavior is hard to learn using reinforcement learning, given
sparsity in reward. This leads to excessively large sample complexity for learning. 
To overcome this large sample complexity, we pre-train our policy to imitate
human demonstrations of how to explore a new environment. We do this using 
trajectories collected from AMT workers as they were answering questions in 
House3D~\citep{embodiedqa}. We ignore the question that was posed to the human,
and treat these trajectories as a proxy for how a human will explore a previosuly unseen
environment. After this phase of imitation learning, $\pi_e$ is further trained  
via policy gradients \citep{williams1992simple} using proximal policy 
optimization (PPO) from \cite{schulman2017proximal}. 
\section{Experiments}
The goal of this paper is to build agents that can autonomously
explore novel complex 3D environments. We want to understand this in context 
of the different choices we made in our design, as well as how our design compares to
alternate existing techniques for exploration. While coverage
of the novel environment is a good task-agnostic metric, we also design experiments
to additionally quantify the utility of generic task-agnostic exploration 
for downstream tasks of interest. We first describe our experimental setup that 
consists of complex realistic 3D environments and emphasizes the study of generalization
to novel environments. We then describe experiments that measure task-agnostic exploration
via coverage. And finally, we present experiments where we use different exploration schemes
for downstream navigation tasks.

\subsection{Experimental Setup}
We conducted our experiments on the House3D simulation environment~\cite{wu2018building}.
\house is based on realistic apartment layouts from the SUNCG dataset~\cite{song2017semantic}
and simulates first-person observations and actions of a robotic agent embodied in these 
apartments. We use 20 houses each for training and testing. These sets are sampled
from the respective sets in \house, and do not overlap. That is, testing is done on
a set of houses not seen during training. This allows us to study generalization, \ie, how well 
our learned policies perform in novel, previously unseen houses. We made one customization 
to the \house environment: by default doors in \house are rendered but not used for collision checking. We 
modified \house to also not render doors, in addition to not using them for collision
checking.

\textbf{Observation Space.}
We assume that the agent has an \rgbd camera with a field of view of $60^\circ$, 
and a bump sensor. The \rgbd camera returns a regular \rgb image and a depth image 
that records the distance (depth information is clipped at $3\si{\meter}$) of each pixel from the camera.  The bump sensor returns if a collision happened while executing the previous action.

\textbf{Action Space.}
We followed the same action space as in \embodiedqa~\citep{embodiedqa}, 
with $6$ motion primitives: \textit{move forward} $0.25\si{\meter}$, 
\textit{move backward} $0.25\si{\meter}$, \textit{strafe left} $0.25\si{\meter}$, 
\textit{strafe right} $0.25\si{\meter}$, \textit{turn left} $9\si{\degree}$, 
and \textit{turn right} $9\si{\degree}$.

\textbf{Extrinsic Environmental Reward.}
We do not use any externally specified reward signals from the environment: $R_{ext}(t) = 0$.

\textbf{Intrinsic Reward.}
As described in \secref{sec:reward}, the intrinsic reward of the agent is based upon the map that it constructs
as it moves around (as described in \secref{sec:make-map}), and readings from the bump
sensor. Reward $R_{int}(t) = \alpha R_{int}^{\text{cov}}(t) + \beta R_{int}^{\text{coll}}(t)$. 
Here $R_{int}^{\text{cov}}$ is the coverage reward, and $R_{int}^{\text{coll}}$ is the collision reward.
$\alpha, \beta$ are hyper-parameters to trade-off how aggressive the agent is.

\textbf{Training.}
As described in \Secref{sec:training}, we train policies using imitation of human trajectories
and RL.
\begin{enumerate}[leftmargin=*]
\item \textit{Imitation from Human Exploration Trajectories.}
    We leverage human exploration trajectories collected from AMT workers as they 
    answered questions for the \embodiedqa task \citep{embodiedqa}. We ignore the question that
    was posed to the AMT worker, and train our policy to simply mimic the actions that the humans took.
    As the humans were trying to answer these questions in previously unseen environments,
    we assume that these trajectories largely exhibit exploration behavior. We used a total
    of $693$ human trajectories for this imitation.

\item \textit{Reinforcement Learning}: After learning via imitation, we further train the policy
    using reinforcement learning on the training houses. We use PPO \citep{schulman2017proximal}
    to optimize the intrinsic reward defined above. At the start of each episode, the agent 
    is initialized at a random location inside one of the training houses. Each episode is run for
    500 time-steps. We run a total of 6400 episodes which amounts to a total of 3.2M steps
    of experience.
\end{enumerate}

\textbf{Baselines.} We next describe the various baseline methods that we experimented with.
We implemented a classical baseline that purely reasons using geometry, and a learning
baseline that uses curiosity for exploration.
\begin{enumerate}[leftmargin=*]
\item \textit{Frontier-based Exploration.} As a classical baseline, we 
    experimented with frontier-based exploration~\citep{yamauchi1997frontier, dornhege2013frontier}. This is a purely geometric 
    method that utilizes the built map $M_t$. Every iteration, it samples a point
    in currently unexplored space, and plans a path towards it from the current
    location (unobserved space is assumed to be free). As the plan is executed, both the map and the plan are updated.
    Once the chosen point is reached, this process is repeated.
\item \textit{Curiosity-based Exploration.} The next baseline we tried was 
    curiosity-based exploration. In particular, we use the version proposed by 
    \citet{pathak2017curiosity} that uses prediction error of a forward model as reward.
    We use the modifications proposed by \citet{burda2018large}, and only train
    a forward model. We prevent degeneracy in forward model by learning it in 
    the fixed feature space of a ResNet-18 model that has been pre-trained on the ImageNet
    classification task.
\end{enumerate}
\subsection{Coverage Quality}
\label{subsec:cov_qua}
We first measure exploration by itself, by measuring the \textit{true}
coverage of the agent. We compute the true coverage using the map as described 
in \secref{sec:make-map}, except for using the true location of the
agent rather than the estimated location (\ie $x_t$ instead of $\hat{x}_{t}$).
We study the following three scenarios:

\begin{figure}
\centering
\vspace{-30pt}
\insertW{0.3}{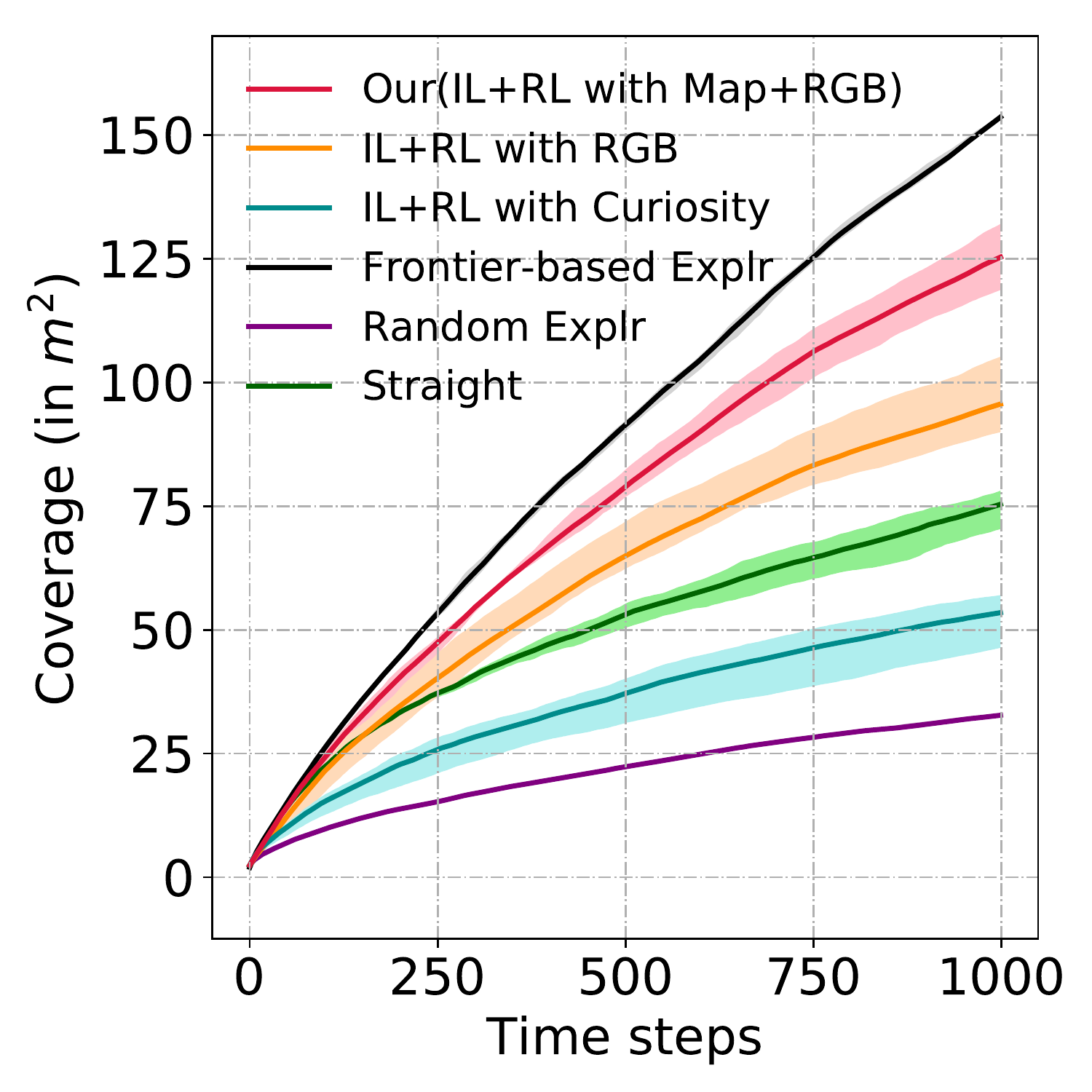}
\insertW{0.3}{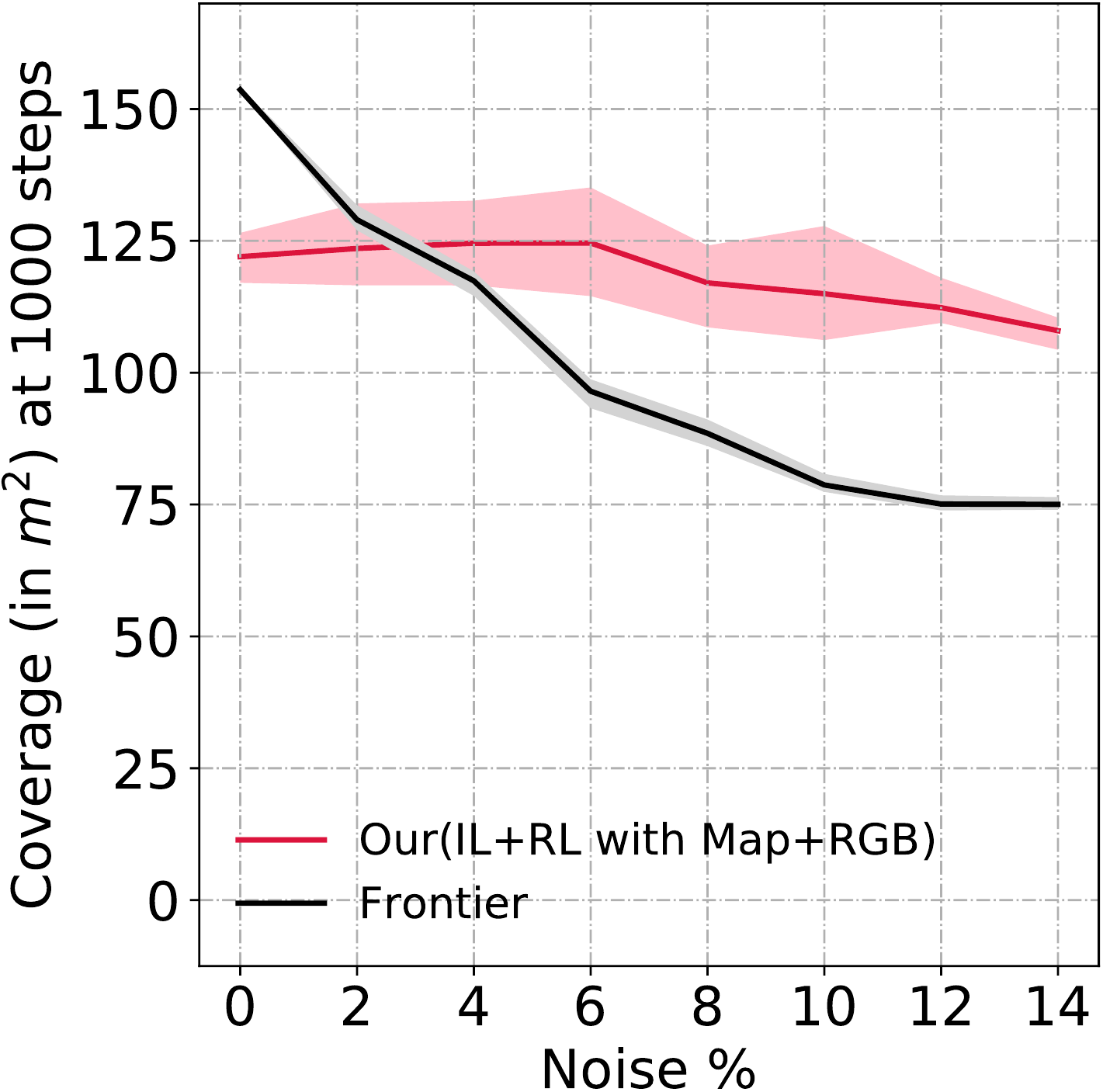}
\insertW{0.3}{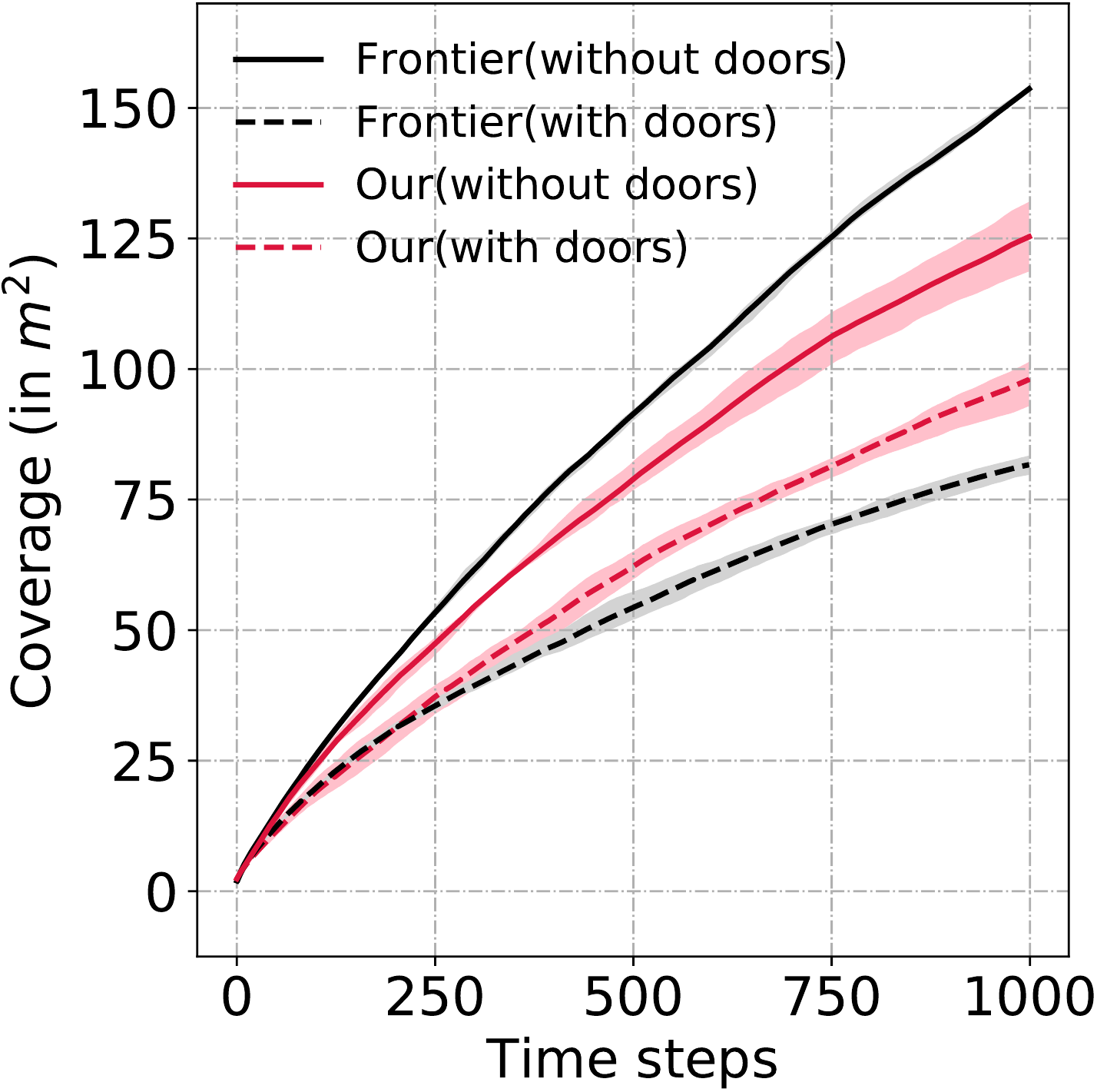}
\caption{\textbf{Coverage Performance.}
Policies are tested on 100 exploration runs (5 random start locations each on 20 testing houses).
\textbf{Left figure} plots average coverage as a function of number of time steps 
in the episode. \textbf{Center figure} studies impact of noise in state estimation.
We plot average coverage at end of episode as a function of amount of noise. \textbf{Right figure}
studies the case when there is a mis-match between geometry and affordance, and plots 
average coverage as a function of time steps in episode. All plots report mean 
performance over $3$ runs, and the shaded area represents the minimum and maximum performance.} 
\label{fig:coverage}
\end{figure}

 \textbf{Without Estimation Noise:} We first
    evaluate the performance of the agent that is trained and tested without
    observation noise, that is, $\hat{x}_t = x_t$. Note that \textit{this setting
    is not very realistic as there is always observation error in an agent's estimate of its location.}
    It is also highly favorable to the frontier-based exploration agent, 
    which very heavily relies on the accuracy of its maps. 
    \Figref{fig:coverage}(left) presents the performance of different 
    policies for this scenario.
    We first note that the curiosity based agent (\textbf{IL + RL with Curiosity})
    explores better than a random exploration policy 
    (that executes a random action at each time step). \textbf{Straight} is a baseline where the agent moves
    along a straight line and executes a random number of $9^\circ$ turns when a collision occurs, which is a strategy used by many robot vacuums.
    Such strategy does not require \rgb or depth information, and performs better than the curiosity based agent.
    However, both policies are worse than an \rgb only version 
    of our method, an \rgb only policy that is trained with our coverage 
    reward (\textbf{IL + RL with \rgb}). Our full system 
    (\textbf{IL + RL with Map + \rgb}) that also uses maps as input performs even better.
    Frontier-based exploration (\textbf{Frontier-based Explr}) has the best performance in this scenario. 
    As noted, this is to expect as this method gets access to perfect, fully-registered 
    maps, and employs optimal path planning algorithm to move from one place to another.
    It is also worth noting that, it is hard to use such classical 
    techniques in situations where we only have a \rgb images. In contrast, learning
    allows us to easily arrive at policies that can explore using only
    \rgb images at test time.
    
\textbf{With Estimation Noise:} We next describe a more realistic scenario that 
    has estimation noise, \ie $\hat{x}_{t+1}$ is estimated using a noisy
    dynamics function. In particular, we add truncated Gaussian noise to the transition 
    function $f$ at each time step. The details of the noise generation is elaborated in Appendix \ref{app:noise_model}. The noise compounds over time. Even though such a noise model leads to compounding errors over time (as in the case of a real robot), we acknowledge that this simple noise model may not perfectly match noise in the real world.
    \Figref{fig:coverage}(center) presents the coverage area at the end of
    episode (1000 time steps) as a function of the amount of noise 
    introduced\footnote{A noise of $\eta$ means we sample perturbations from a truncated Gaussian distribution with zero-mean,  
    $\eta$ standard deviation, and a 
    total width of $2\eta$. This sampled perturbation is scaled by the step length
    (25cm for $x, y$ and $9^\circ$ for azimuth $\theta$) and added to the state at
    each time step.}.
    When the system suffers from sensor noise, the performance of 
    the frontier-based exploration method drops rapidly. In contrast,
    our learning-based agent that wasn't even trained with any noise continues 
    to performs well. Even at relatively modest noise of 4\% 
    our learning based method already does better than the frontier-based agent.
    We additionally note that our agent can even be trained when the intrinsic
    reward estimation itself suffers from state estimation noise: for instance
    performance with 10\% estimation noise (for intrinsic reward computation and
    map construction) is $98.9m^2$, only a minor degradation from $117.4m^2$ (10\% 
    estimation noise for map construction at test time only).
    
\textbf{Geometry and Affordance Mismatch:} Next, to emphasize the
    utility of learning for this task, we experiment with a scenario
    where we explicitly create a mismatch between geometry and affordance of the 
    environment. We do this by rendering doors, but not using them for
    collision checking (\ie the default \house environment).
    This setting helps us investigate if learning based techniques go beyond
    simple geometric reasoning in any way. \Figref{fig:coverage}(right) 
    presents performance curves. We see that there is a large drop in
    performance for the frontier-based agent. This is because it is not
    able to distinguish doors from other obstacles, leading to path planning
    failures. However, there is a relatively minor drop in performance of our
    learning-based agent. This is because it can learn about doors (how to 
    identify them in \rgb images and the fact that they can be traversed) 
    from human demonstrations and experience during reinforcement learning.

\subsubsection{Ablation Study}
We also conducted ablations of our method to identify what parts of our
proposed technique contribute to the performance. We do these ablations
in the setting without any estimation noise, and use coverage as the metric.

\begin{figure}
\centering 
\vspace{-30pt}
\insertW{0.3}{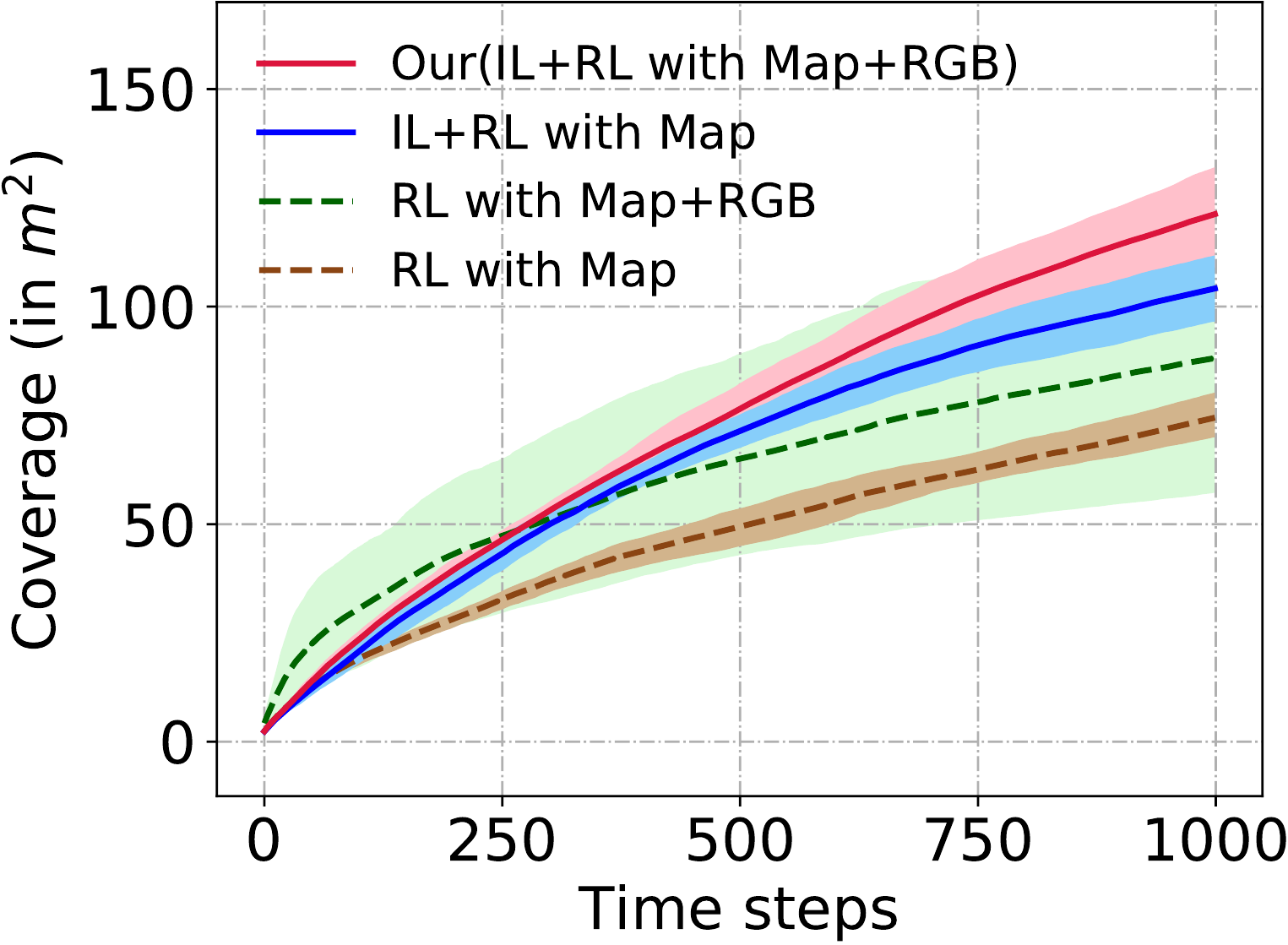}
\insertW{0.3}{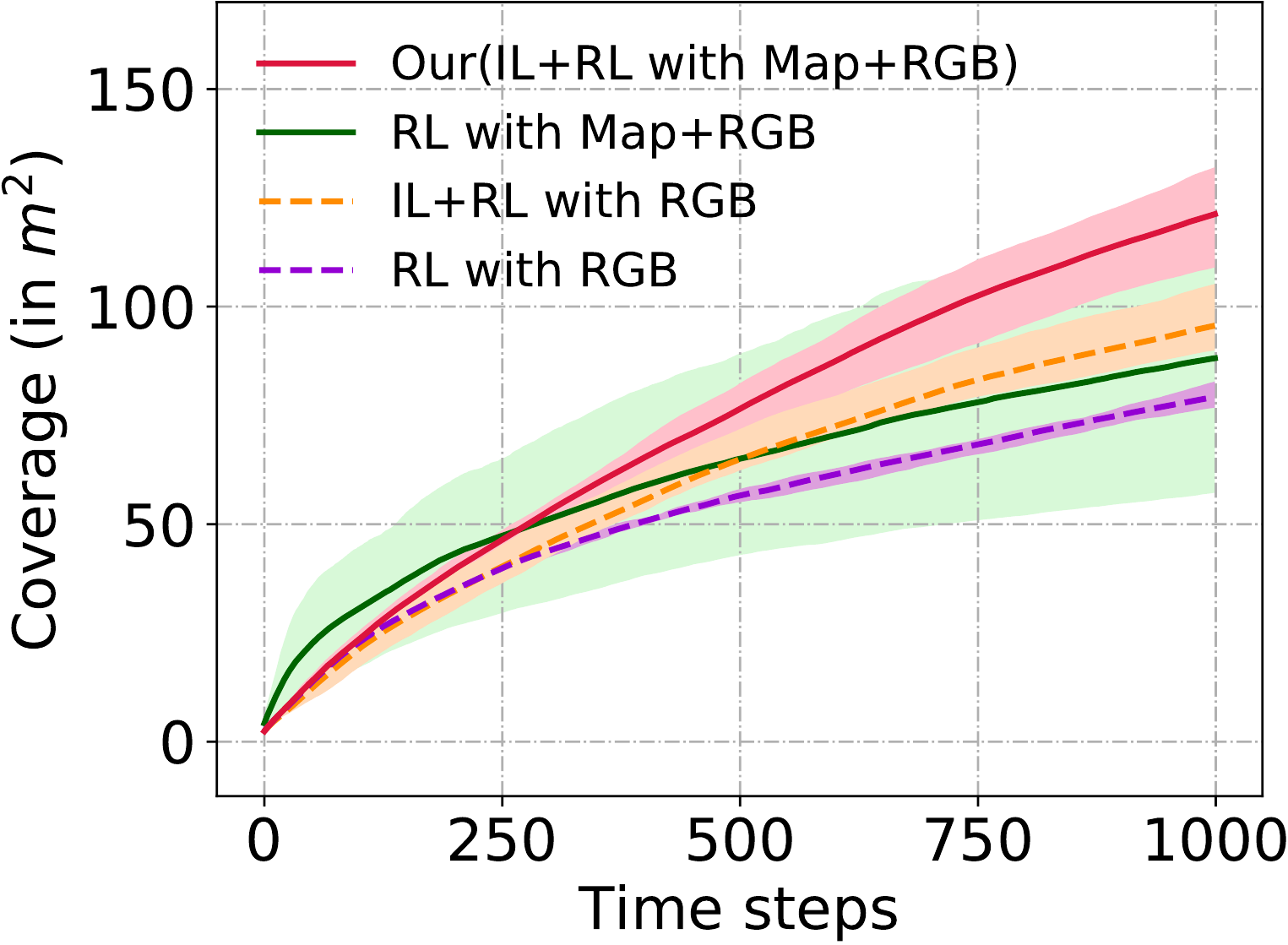}
\insertW{0.3}{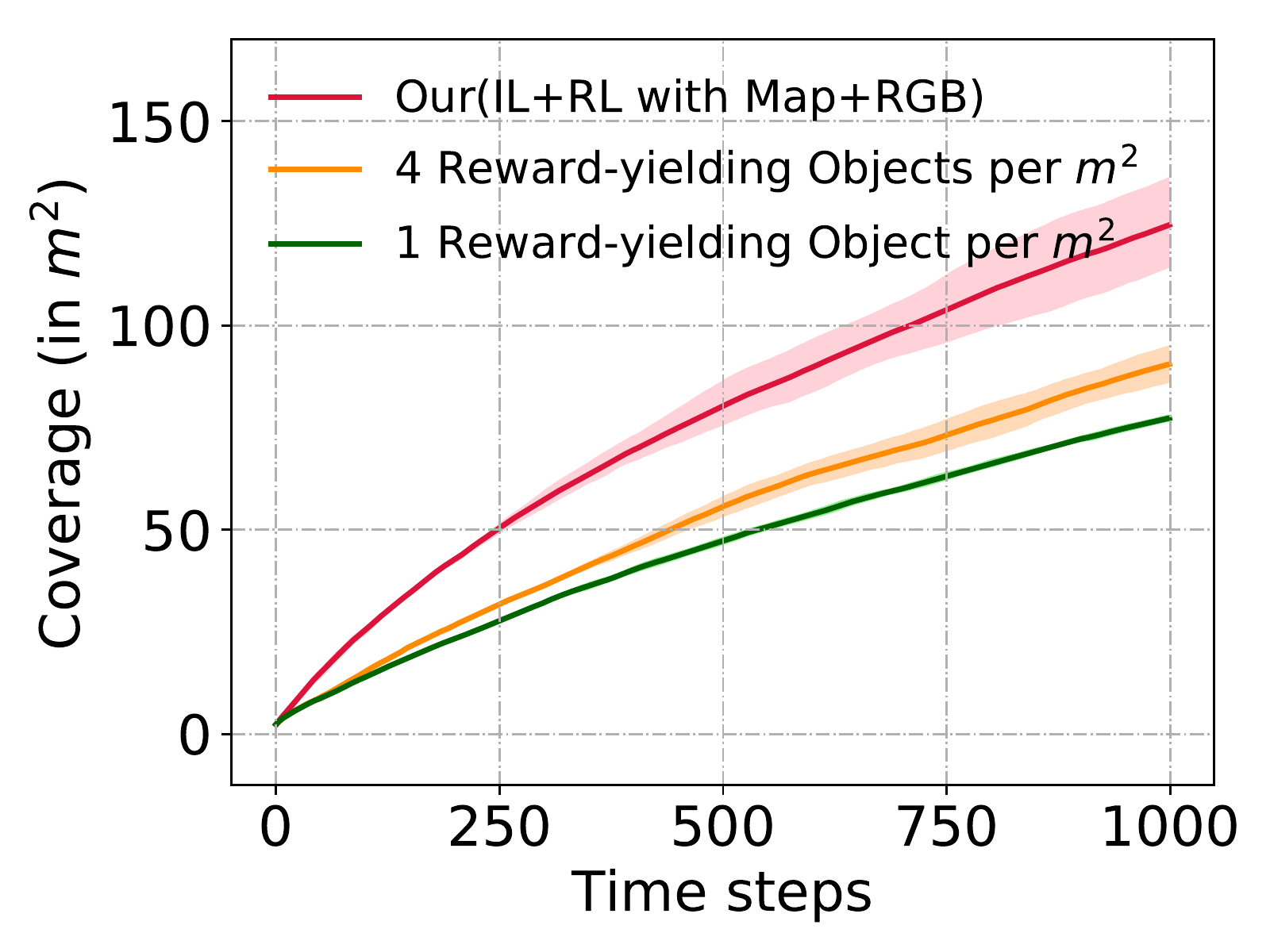}
\caption{\textbf{Ablation Study:} We report average coverage
as a function of time step in episode. As before we plot
mean over 3 runs and minimum and maximum performance.
\textbf{Left figure} shows that using the \rgb image 
helps improve performance, \textbf{center figure} shows that 
using the map helps improve performance. We can also see that imitation 
learning improves coverage and reduces the variance in performance. \textbf{Right figure} shows the comparison between different reward design. We can see that our intrinsic reward enables the agent to explore more efficiently in the testing time.}
\label{fig:ablation}
\vspace{-15pt}
\end{figure}

\textbf{Imitation Learning:} We check if pre-training with 
    imitation learning is useful for this task. We test this by 
    comparing to the models that were trained only with RL 
    using the coverage reward. The left two plots in \Figref{fig:ablation}
    shows performance of agents with following combinations: 
    \textit{RL}: policy trained with PPO only, 
    \textit{IL}: policy trained with imitation learning, 
    \textit{Map}: policy uses constructed maps as input, 
    \textit{RGB}: policy uses RGB images as input. 
    In all settings, pre-training with imitation learning helps improve performance, 
    though training with RL improves coverage further.
    Also, policies trained using RL have a fairly large variance. 
    Imitation learning also helps reduce the variance.

\textbf{\rgb Observations and Map:} \Figref{fig:ablation}~(left) and \Figref{fig:ablation}~(center)
    respectively show that both \rgb images and map inputs improve performance.
    
\textbf{Intrinsic Reward:}  We also compare our intrinsic reward design with extrinsic reward design, 
as shown in \Figref{fig:ablation}~(right). The extrinsic reward is setup as follows: we randomly generated
a number of locations evenly distributed across the traversable area of the houses, where the agent will get
a positive reward if the agent is close to any of these locations. Once the agent gets a reward from a location, 
this location will be no longer taken into account for future reward calculation.
We tried two settings where we place 1 or 4 reward-yielding objects per $m^2$.
We can see that our coverage map reward provides a better reward signal and in fact can be estimated
intrinsically without needing to instrument the environment.

\subsection{Using Exploration for Downstream Task}
\setlength{\intextsep}{0pt}
\setlength{\columnsep}{6pt}%
\begin{wrapfigure}[23]{r}{0.35\textwidth}
    \begin{center}
      \insertW{0.35}{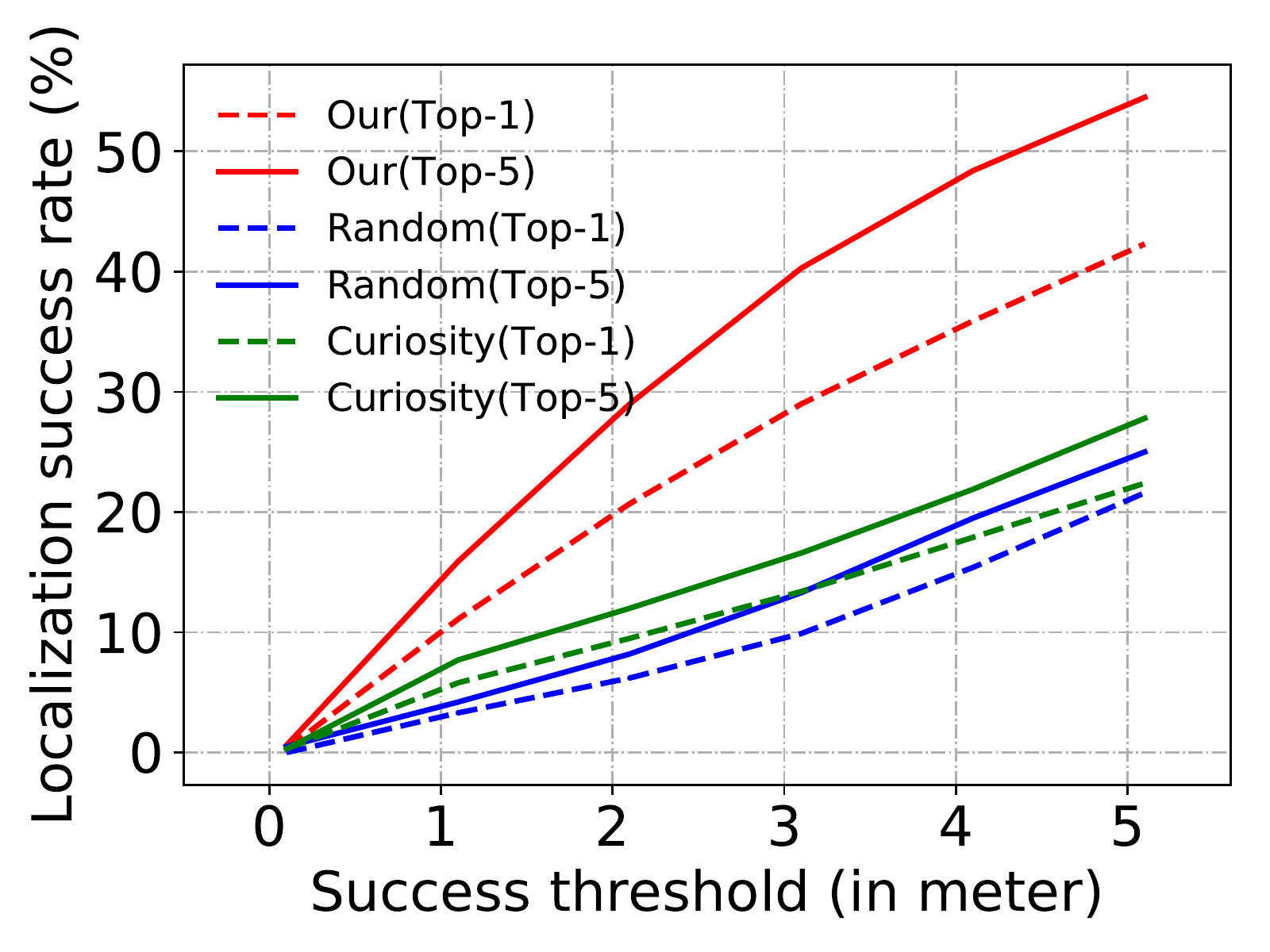} \\ \vspace{6pt}
      \resizebox{1.0\linewidth}{!}{
        \setlength{\tabcolsep}{6pt}
        \begin{tabular}{ccc} \toprule
SPL (No exploration) &  SPL (Our exploration) \\ \midrule
0.69                 & \textbf{0.87}         \\ \bottomrule
        \end{tabular}}
    \end{center}
    \caption{\textbf{Exploration for downstream tasks.} We evaluate utility of exploration
    for the down-stream navigation tasks. \textbf{Top plots} shows effectiveness of
    the exploration trajectory for localization of goal images in a new environment.
    \textbf{Bottom table} compares efficiency of reaching goals without and with maps 
    built during exploration.}
    \label{fig:spl}
\end{wrapfigure}
Now that we have established how to explore well, we next ask if
task-agnostic exploration is even useful for downstream tasks.
We show this in context of goal-driven navigation.
We execute the learned exploration policy $\pi_e$ (for $1500$ time steps)
to explore a new house, and collect experience (set of images $\mathcal{I}$ 
along with their pose $\bf{p}$).
Once this exploration has finished, the agent is given
different navigation tasks. These navigation tasks put the 
agent at an arbitrary location in the environment, and give it a \textit{goal image} 
that it needs to reach. More specifically, we reset the agent to $50$ random  poses (positions and orientations)
in each testing house (20 testing houses in total) and get the RGB camera view. 
The agent then needs to use the experience of the
environment acquired during exploration to efficiently navigate to the desired
target location. The efficiency of navigation on these test queries measures
the utility of exploration.

This evaluation requires a navigation policy $\pi_n$, that uses 
the exploration experience and the goal image to output
actions that can convey the agent to the desired target location. We opt for 
a simple policy $\pi_n$. $\pi_n$ first localizes the target image using 
nearest neighbor matching to the set of collected images $\mathcal{I}$ 
(in ImageNet pre-trained ResNet-18 feature space). It then plans a path 
to the this estimated target location using a occupancy map
computed from $\mathcal{I}$. We do this experiment in the setting without
any state estimation noise.

We independently measure the effectiveness of exploration data for a) localization of
the given target images, and b) path planning efficiency in reaching desired locations. 
\textbf{Localization performance:} We measure the distance between the agent's estimate
of the goal image location, and the true goal image location. 
\Figref{fig:spl}~(top) plots the success at localization as a function of the 
success threshold (distance at which we consider a localization as correct).
We report top-1 and top-5 success rates. We compare to the random exploration 
baseline and curiosity-driven baseline which serve to measure the hardness of the task.
We see our exploration 
scheme performs well.
\textbf{Path planning efficiency:} Next, we measure
how efficiently desired goal locations can be reached. We measure performance
using the \spl metric as described by \citet{anderson2018evaluation} 
(described in the appendix, higher is better). We compare against
a baseline that does not have any prior experience in this environment and derives
all map information on the fly from goal driven behavior (going to the desired test location).
Both agents take actions based on the shortest-path motion planning algorithm. 
Once again these serve as a measure of the hardness of the topology of the underlying
environment. As shown in \Figref{fig:spl}~(bottom), using exploration 
data from our policy improves efficiency of paths to reach target locations.
\section{Discussion}
In this paper, we motivated the need for learning explorations policies
for navigation in novel 3D environments. We showed how to design 
and train such policies, and how experience gathered from such policies enables better performance for 
downstream tasks. We think we have just scratched the surface of this problem,
and hope this inspires future research towards semantic exploration using 
more expressive policy architectures, reward functions and training techniques.

\bibliography{refs}
\bibliographystyle{iclr2019_conference}
\newpage
\renewcommand{\thefigure}{A\arabic{figure}}
\setcounter{figure}{0}
\begin{appendices}
\renewcommand{\thefigure}{\Alph{section}.\arabic{figure}}
\renewcommand{\thetable}{\Alph{section}.\arabic{table}}
\section{SPL Metric}
As defined by \cite{anderson2018evaluation}, Success weighted by (normalized inverse) 
Path Length or SPL is computed as follows. 
Here, $\ell_i$ is the shortest-path distance between the starting and goal position,
$p_i$ is the length of the path actually executed by the agent, and $S_i$ is a binary
variable that indicates if the agent succeeded. SPL is computed over $N$ trials as follows:
\begin{equation}
  SPL = \frac{1}{N} \sum_{i=1}^{N} S_i \frac{\ell_i}{\max(p_i,\ell_i)}.
\end{equation}

\section{Examples of Policy Inputs and Maps}
\begin{figure}[H]
\centering
\begin{subfigure}{0.28\textwidth}
  \centering
  \includegraphics[width=0.99\linewidth]{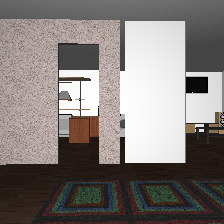}
  \caption{}
  \label{fig:r1}
\end{subfigure}%
\begin{subfigure}{0.28\textwidth}
  \centering
  \includegraphics[width=0.99\linewidth]{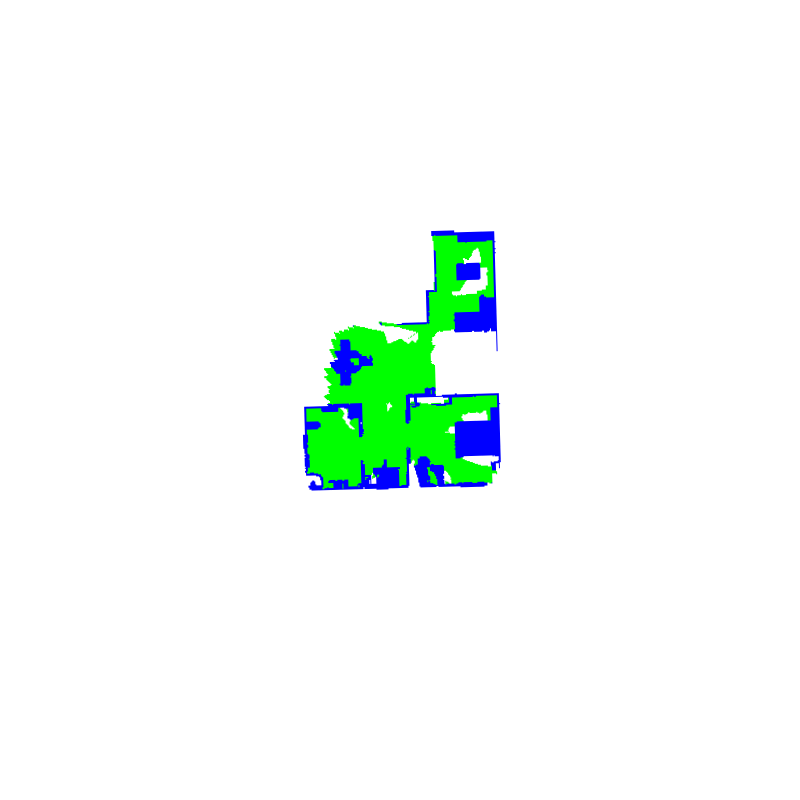}
  \caption{}
  \label{fig:l1}
\end{subfigure}%
\begin{subfigure}{0.28\textwidth}
  \centering
  \includegraphics[width=0.99\linewidth]{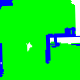}
  \caption{}
  \label{fig:s1}
\end{subfigure}
\begin{subfigure}{0.28\textwidth}
  \centering
  \includegraphics[width=0.99\linewidth]{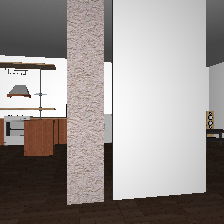}
  \caption{}
  \label{fig:r2}
\end{subfigure}%
\begin{subfigure}{0.28\textwidth}
  \centering
  \includegraphics[width=0.99\linewidth]{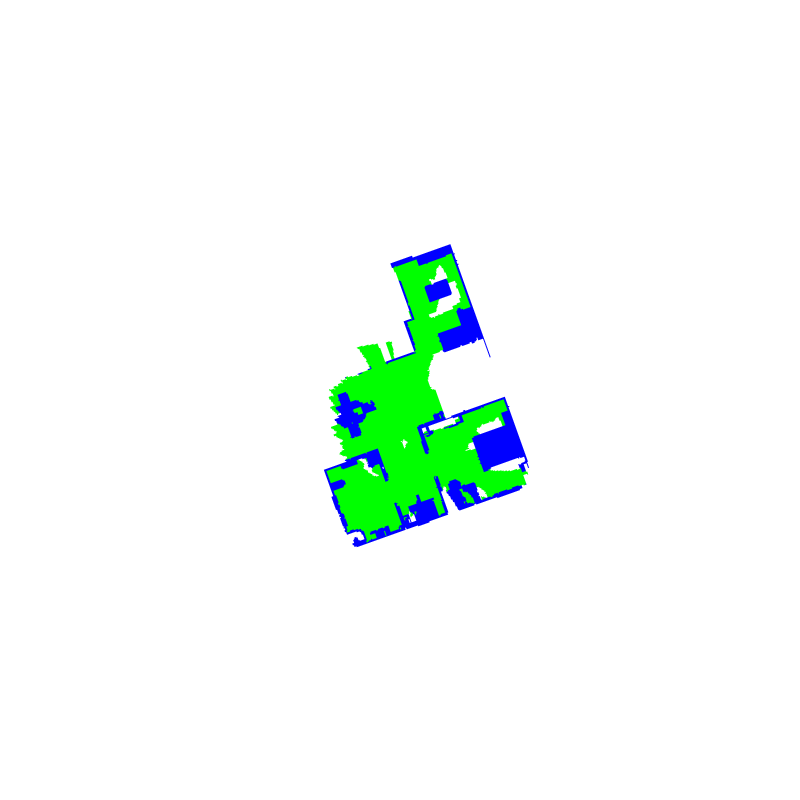}
  \caption{}
  \label{fig:l2}
\end{subfigure}%
\begin{subfigure}{0.28\textwidth}
  \centering
  \includegraphics[width=0.99\linewidth]{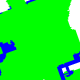}
  \caption{}
  \label{fig:s2}
\end{subfigure}
\caption{Examples of policy inputs at time step $t$ and $t+10$. (a) and (d) are the RGB images, (b) and (e) are the coarse maps, and (c) and (f) are the fine maps.}
\label{fig:policy_input}
\end{figure}

\begin{figure}[H]
\centering
\begin{subfigure}{0.16\textwidth}
  \centering
  \includegraphics[width=0.99\linewidth, frame]{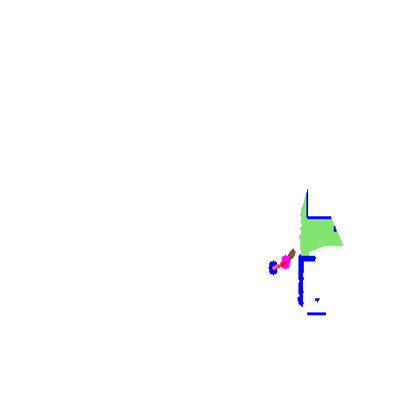}
  \caption{}
  \label{fig:m1}
\end{subfigure}%
\begin{subfigure}{0.16\textwidth}
  \centering
  \includegraphics[width=0.99\linewidth, frame]{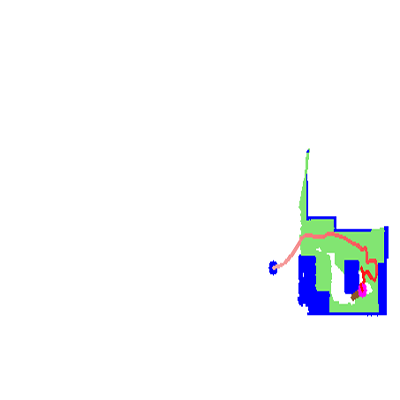}
  \caption{}
  \label{fig:m2}
\end{subfigure}%
\begin{subfigure}{0.16\textwidth}
  \centering
  \includegraphics[width=0.99\linewidth, frame]{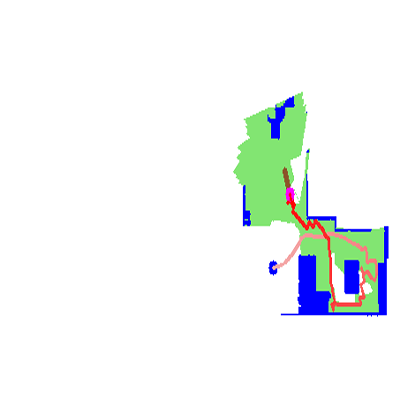}
  \caption{}
  \label{fig:m3}
\end{subfigure}%
\begin{subfigure}{0.16\textwidth}
  \centering
  \includegraphics[width=0.99\linewidth, frame]{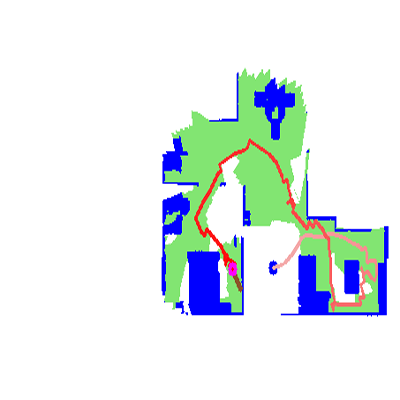}
  \caption{}
  \label{fig:m4}
\end{subfigure}%
\begin{subfigure}{0.16\textwidth}
  \centering
  \includegraphics[width=0.99\linewidth, frame]{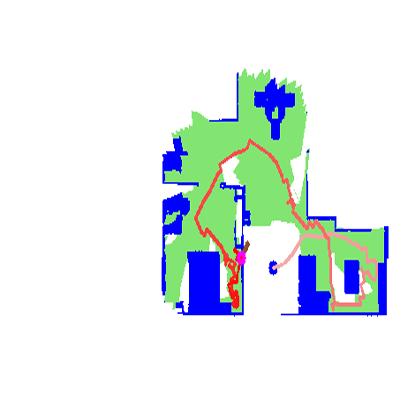}
  \caption{}
  \label{fig:m5}
\end{subfigure}%
\begin{subfigure}{0.16\textwidth}
  \centering
  \includegraphics[width=0.99\linewidth, frame]{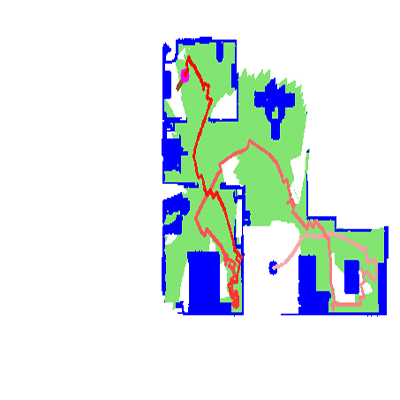}
  \caption{}
  \label{fig:m6}
\end{subfigure}
\begin{subfigure}{0.16\textwidth}
  \centering
  \includegraphics[width=0.99\linewidth, frame]{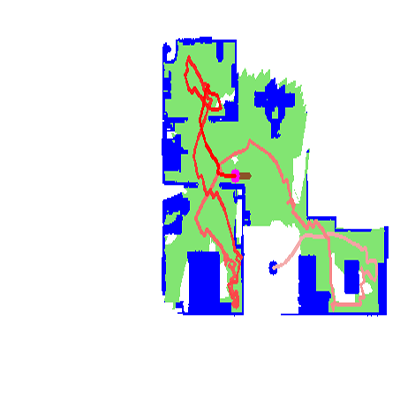}
  \caption{}
  \label{fig:m7}
\end{subfigure}%
\begin{subfigure}{0.16\textwidth}
  \centering
  \includegraphics[width=0.99\linewidth, frame]{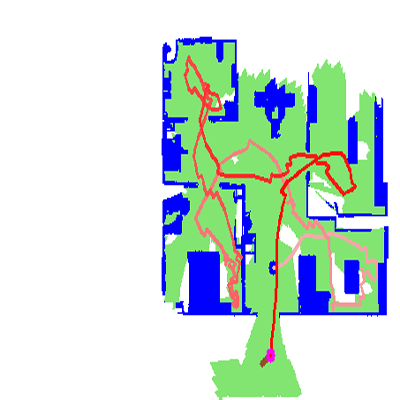}
  \caption{}
  \label{fig:m8}
\end{subfigure}%
\begin{subfigure}{0.16\textwidth}
  \centering
  \includegraphics[width=0.99\linewidth, frame]{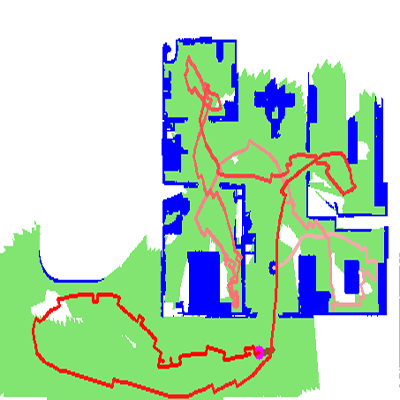}
  \caption{}
  \label{fig:m9}
\end{subfigure}%
\begin{subfigure}{0.16\textwidth}
  \centering
  \includegraphics[width=0.99\linewidth, frame]{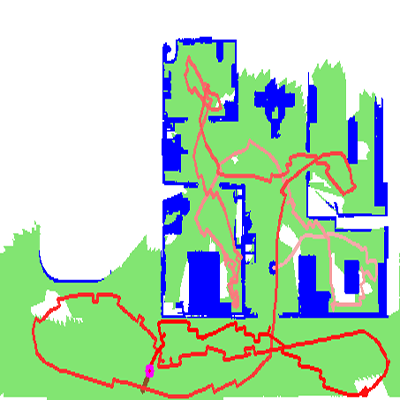}
  \caption{}
  \label{fig:m10}
\end{subfigure}%
\begin{subfigure}{0.16\textwidth}
  \centering
  \includegraphics[width=0.99\linewidth, frame]{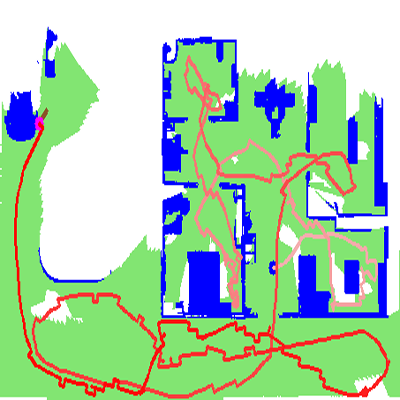}
  \caption{}
  \label{fig:m11}
\end{subfigure}%
\begin{subfigure}{0.16\textwidth}
  \centering
  \includegraphics[width=0.99\linewidth, frame]{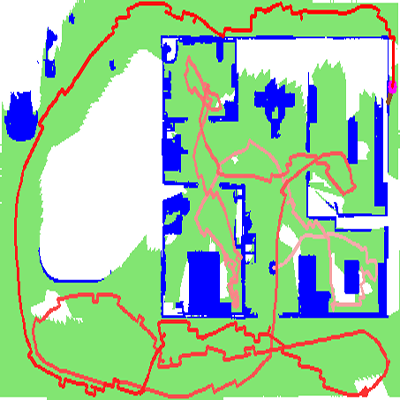}
  \caption{}
  \label{fig:m12}
\end{subfigure}%
\caption{Snapshots of the built map as the agent explores the house }
\label{fig:map}
\end{figure}

We show an example of the policy inputs (RGB image and 
maps in two different scales) at time $t$ and $t+10$ in \figref{fig:policy_input}. Green area means the known(seen) traversable area, blue area means known non-traversable area, and the white area means unknown area.
\figref{fig:map} shows how the map evolves and the agent moves as it explores a new house.

\section{Experimental Details}

\setcounter{figure}{0}
\subsection{Environment Details}
\begin{figure}[H]
    \centering
    \includegraphics[width=0.4\linewidth]{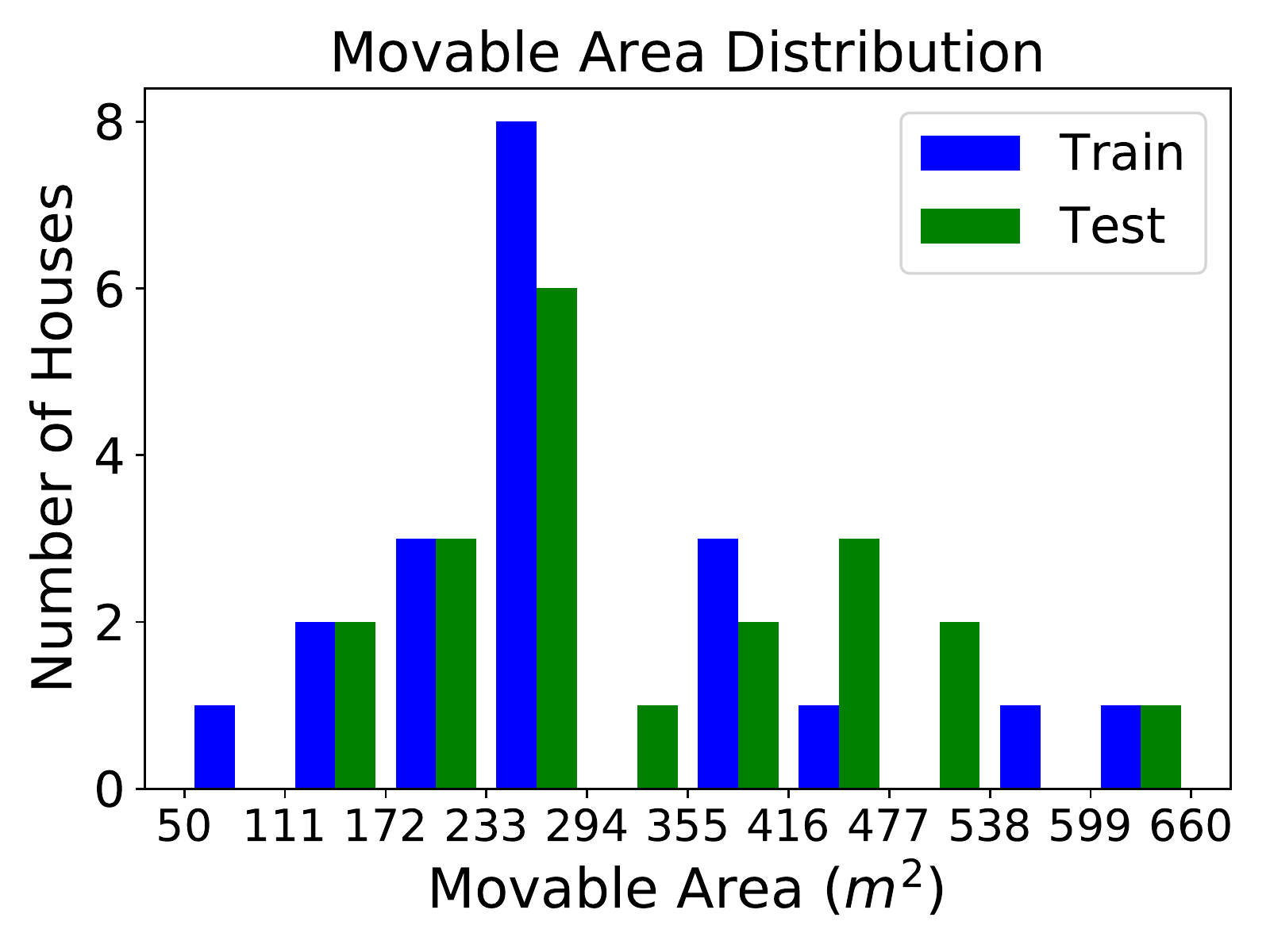}
    \caption{Area distribution of training and testing houses (20 houses in each case). The average movable area of the training houses is $289.7$ m$^2$. The average movable area of the testing houses is $327.9$ m$^2$.}
    \label{fig:avg_area}
\end{figure}

\begin{figure}[H]
\centering
\begin{subfigure}{0.16\textwidth}
  \centering
  \includegraphics[width=0.9\linewidth, frame]{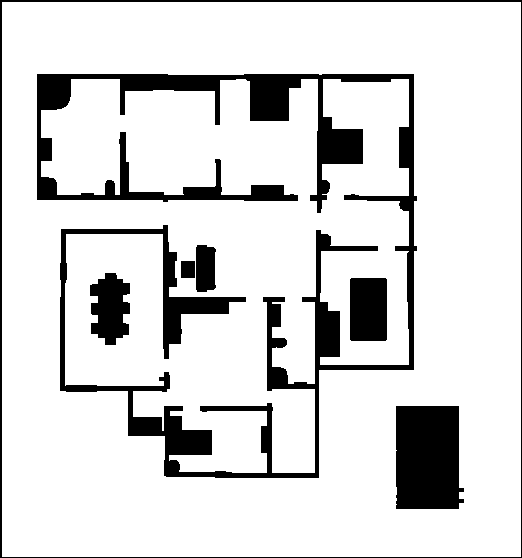}
\end{subfigure}%
\begin{subfigure}{0.16\textwidth}
  \centering
  \includegraphics[width=0.9\linewidth, frame]{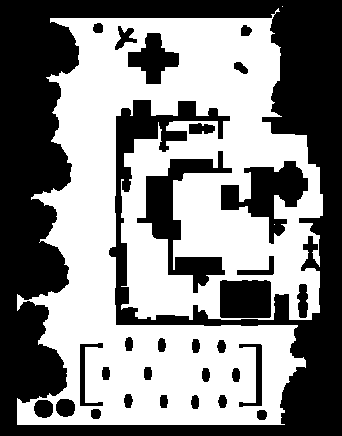}
\end{subfigure}%
\begin{subfigure}{0.16\textwidth}
  \centering
  \includegraphics[width=0.9\linewidth, frame]{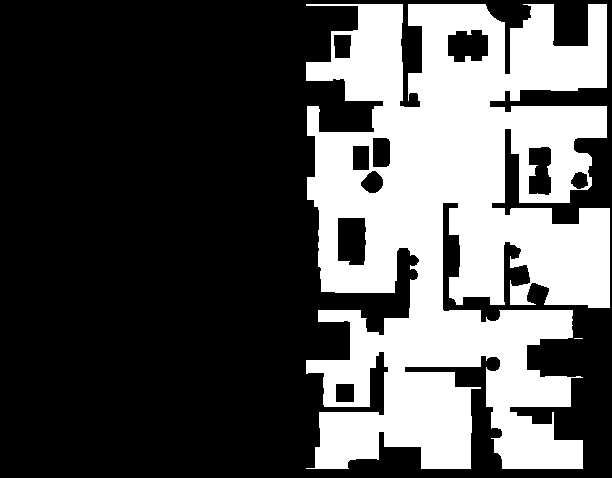}
\end{subfigure}
\unskip\ \vrule\ 
\begin{subfigure}{0.16\textwidth}
  \centering
  \includegraphics[width=0.9\linewidth, frame]{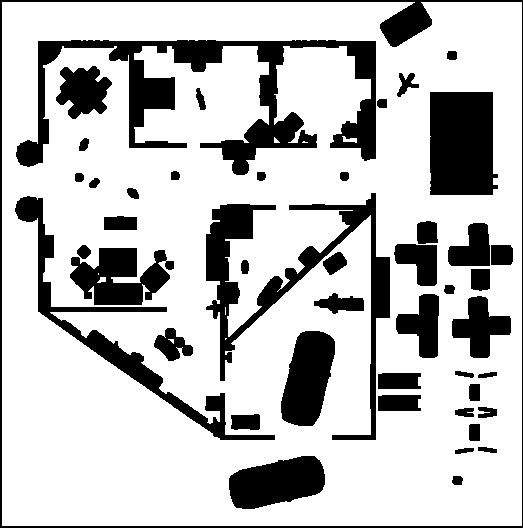}
\end{subfigure}%
\begin{subfigure}{0.16\textwidth}
  \centering
  \includegraphics[width=0.9\linewidth, frame]{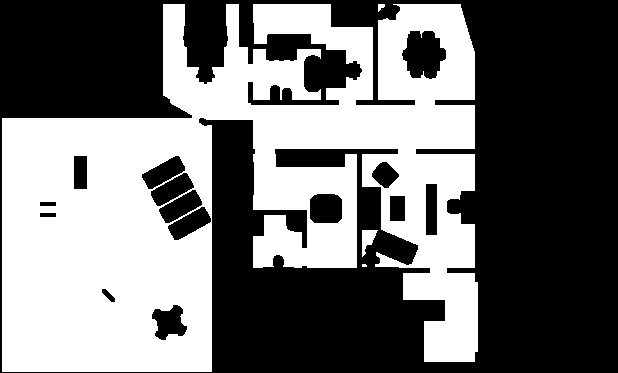}
\end{subfigure}%
\begin{subfigure}{0.16\textwidth}
  \centering
  \includegraphics[width=0.9\linewidth, frame]{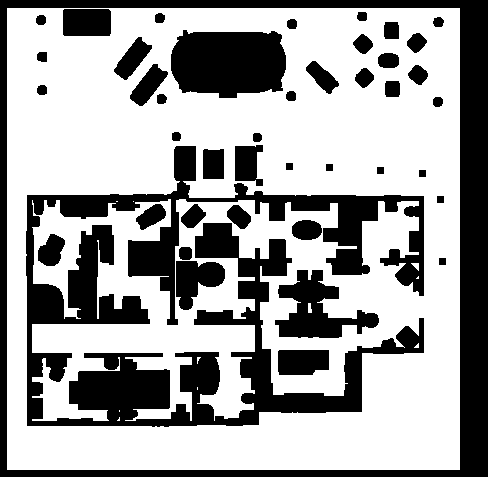}
\end{subfigure}
\caption{Examples of house layouts (top-down views). \textbf{The left} three figures show $3$ examples of the training houses. \textbf{The right} three figures show $3$ examples of the testing houses. White regions indicate traversable area, black regions represent occupied area.}
\label{fig:layout}
\end{figure}

In this paper, we simulated the agent in House3D 
environment \cite{wu2018building} and used $20$ houses 
for training and $20$ new houses for testing.
\figref{fig:avg_area} shows the distribution of the 
total traversable area of these houses.
\figref{fig:layout} shows some examples of the top-down views for 
training and testing houses (white is traversable, black is occupied).

\subsection{Map Reconstruction}
Given a sequence of camera poses (extrinsics) ($[\boldsymbol{R}_1, \boldsymbol{t}_1], [\boldsymbol{R}_2, \boldsymbol{t}_2], ..., [\boldsymbol{R}_N, \boldsymbol{t}_N]$), and the corresponding depth images ($\boldsymbol{D}_1, \boldsymbol{D}_2, ..., \boldsymbol{D}_N$) at each time step as well as the camera intrinsics $ \boldsymbol{K}\in\mathbb{R}^{3\times 3}$, we can reconstruct the point cloud of the scene based on the principle of mulitple-view geometry \citep{hartley2003multiple}. We can convert the points from pixel space to the camera space and then to the world space. More specifically, the formulas to achieve this can be summarized as follows:
\begin{align}
\label{eqn:proj}
    \lambda_{ij} \boldsymbol{x}_{ij} &= \boldsymbol{K}\left[\boldsymbol{R}_i, \boldsymbol{t}_i\right]\boldsymbol{w}_{ij} \quad \forall j \in \{1, 2, ..., S\}, i \in \{1, ..., N\}\\
\label{eqn:union}
    \boldsymbol{W} &= \bigcup\limits_{i=1}^{N}\bigcup\limits_{j=1}^{S}\boldsymbol{w}_{ij}
\end{align}
where $S$ is the total number of pixels in each depth image, $\boldsymbol{x}_{ij}\in\mathbb{R}^3$ is the homogeneous coordinates for $j_\text{th}$ pixel on $i_\text{th}$ depth image $D_i$, $\lambda_{ij}\in\mathbb{R}$ is the depth value of the $j_\text{th}$ pixel on $i_\text{th}$ depth image $D_i$, and $\boldsymbol{w}_{ij}\in\mathbb{R}^4$ is the homogeneous coordinates for the corresponding point in the world coordinate system. We can get $\boldsymbol{w}_{ij}$ from $\boldsymbol{x}_{ij}$ based on Equation (\ref{eqn:proj}). And we can merge points via Equation (\ref{eqn:union}).

\subsection{Training Details}
The occupancy map generated by the agent itself uses a resolution of $0.05$m. Our policy uses a coarse map, and a detailed map. The coarse map captures information about a $40$m$\times40$m area around the agent, at a resolution of of $0.5m$. The occupancy map is down-sampled from $800 \times 800$ to $80 \times 80$ to generate the coarse map that is fed into the policy network. The detailed map captures information about a $4$m$\times4$m area around the agent at a $0.05m$ resolution. RGB images are rendered at $224\times224$ resolution in House3D and re-sized into $80\times80$ before they are fed into the policy network.

The size of the last fully-connected layer in ResNet18 architecture is modified to 128. Outputs of the three ResNet18 networks are concatenated into a 384-dimensional vector. This is transformed into a 128-dimensional vector via a fully-connected layer. Next, it's fed into a single-layer RNN(GRU) layer with 128 hidden layer size. The output of RNN layer is followed by two heads. One head (the policy head) has two fully-connected layers ($128-32-6$). The other head (the value head) has two fully-connected layers ($128-32-1$). We use ELU as the nonlinear activation function for the fully-connected layers after ResNet18.

Each episode consists of $500$ steps in training. RNN time sequence length is $20$. Coverage area $C(M_t)$ is measured by the number of covered grids in the occupancy map. Coefficient $\alpha$ for the coverage reward $R_{int}^{\text{cov}}(t)$ is $0.0005$, coefficient $\beta$ for $R_{int}^{\text{coll}}(t)$ is $0.006$. PPO entropy loss coefficient is $0.01$. Network is optimized via Adam optimizer with a learning rate of $0.00001$.

\subsection{Noise Model}
\label{app:noise_model}

Details of noise generation for experiments with estimation noise in Section \ref{subsec:cov_qua}: 
\begin{enumerate}[leftmargin=*]
\item Without loss of generality, we initialize the agent at the origin, that is $\hat{x}_0=x_0=\mathbf{0}$.
\item The agent takes an action $a_t$.  We add truncated Gaussian noise to the action primitive(e.g., move forward 0.25m) to get the estimated pose $\hat{x}_{t+1}$,  i.e., $\hat{x}_{t+1} = \hat{x}_t + \tilde{a}_t$  where $\hat{x}_t$ is the estimated pose in time step t and $\tilde{a}_t$ is the action primitive $a_t$ with added noise.
\item Iterate the second step until the maximum number of steps is reached.
\end{enumerate} 
Thus, in this noise model, the agent estimates its new pose based on the estimated pose from the last time step and the executed action. Thus, we don’t use oracle odometry in the noise experiments. This noise model leads to compounding errors over time (as in the case of a real robot), though we acknowledge that this simple noise model may not perfectly match noise in the real world.

\subsection{Imitation Learning Details}
We use behavioral cloning technique \citep{argall2009survey, michalski1998machine, abbott2007behavioral} to imitate the human behaviors in exploring the environments. We got the human demonstration trajectories from \citep{embodiedqa}. We ignored the question-answering part of the data and only used the exploration trajectories. We cleaned up the data by removing the trajectories that have less than $100$ time steps. The trajectories are then converted into short sequences of trajectory segments ($(s_i, a_i, s_{i+1}, a_{i+1}, ..., s_{i+T}, a_{i+T})$, where $T$ is based on the RNN sequence length). The policy is pretrained with behavioral cloning by imitating actions $(a_i, a_{i+1}, ..., a_{i+T})$ from the human demonstrations given the states $(s_i, s_{i+1}, ..., s_{i+T})$.

\end{appendices}
\end{document}